\RequirePackage{amsmath}
\documentclass[runningheads]{llncs}
\usepackage[utf8]{inputenc}
\usepackage[T1]{fontenc}
\usepackage{amsmath,amssymb,mathtools,geometry,hyperref}
\usepackage[normalem]{ulem}
\usepackage{algorithm,algorithmicx,algpseudocode,comment,cleveref}
\usepackage[textsize=scriptsize, 
    disable,
    ]{todonotes}
\usepackage{verbatim}
\usepackage{%
    stmaryrd,
    xspace,
    placeins,
    booktabs,
    tikz-cd
}

\usepackage{amsthm}

\def\orcidID#1{\href{http://orcid.org/#1}{\protect\raisebox{-1.25pt}{\protect\includegraphics{
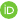%
}}}}

\renewcommand*\vec[1]{\boldsymbol #1}
\newcommand*\todofb[2][]{\todo[color=cyan!30,tickmarkheight=.2em,size=\scriptsize,#1]{FB: #2}}
\newcommand*\todozk[2][]{\todo[color=yellow!30,tickmarkheight=.2em,size=\scriptsize,#1]{ZK: #2}}
\newcommand*\todokk[2][]{\todo[color=purple!20,tickmarkheight=.2em,size=\scriptsize,#1]{KK: #2}}

\newcommand*\KK{\todokk}
\newcommand\changed[1]{{\color{blue}#1}}
\newcommand\removed[1]{}
\newcommand\added[1]{#1}
\newcommand\changedto[2]{%
    \removed{#1}%
    \added{#2}}

\newcommand{\delete}[1]{}
\newcommand*\objv{o}

\newcommand*\dom{\operatorname{dom}}

\newcommand*\query{\operatorname{query}}
\newcommand*\assert{\operatorname{assert}}

\newcommand*\regmin[2]{\llfloor{#1}\rrfloor_{#2}}

\newcommand*\regmax[2]{\llceil{#1}\rrceil_{#2}}

\renewcommand*\implies\to
\newcommand*\feat\textsf

\newcommand*\eqdef=

\title{SMLP: Symbolic Machine Learning Prover%
    \thanks{This research was supported by a grant from Intel Corporation.}}
\author{%
    Franz Brau\ss{}e\inst1\orcidID{0000-0002-2386-7489} \and
    Zurab Khasidashvili\inst2\orcidID{0000-0001-9883-6997} \and
    Konstantin Korovin\inst1\orcidID{0000-0002-0740-621X}}
\institute{%
    The University of Manchester, UK \and
    Intel, Israel}
\begin{document}

\maketitle

\begin{abstract}
\emph{Symbolic Machine Learning Prover (SMLP)} is a tool and a library 
for system exploration
based on data samples obtained by simulating or executing the system on a number of input vectors.
SMLP aims at exploring the system based on this data by taking a grey-box approach:
SMLP combines statistical methods of data exploration with building and exploring machine learning models
in close feedback loop with the system's response,
and exploring these models by combining probabilistic and formal methods.  
SMLP has been applied in industrial setting at Intel for analyzing and optimizing hardware designs at the analog level. SMLP is a general purpose tool and can
be applied to 
systems that can be 
sampled and modeled by machine learning models.




\end{abstract}
\section{Introduction}

Verification of assertions on machine learning (ML) models has received a wide attention from formal methods community in recent years, and multiple approaches have been developed for formal analysis of ML models, mostly focused on neural networks.
In this work we introduce the SMLP tool -- \emph{Symbolic Machine Learning Prover} -- aiming at going beyond this mainstream in several ways:
SMLP
helps to approach the system's design, optimization and verification as one process by offering multiple capabilities for system's \emph{design space exploration}.
These capabilities include methods for selecting which parameters to use in modeling design for configuration optimization and verification;
ensuring that the design is robust against environmental effects and manufacturing variations that are impossible to control, as well as ensuring robustness against malicious attacks from an adversary aiming at altering the intended configuration or mode of operation.
Environmental affects like temperature fluctuation, electromagnetic interference, manufacturing variation, and product aging effects are especially more critical for correct and optimal operation of devices with analog components, which is our current focus.

To address these challenges, SMLP offers multiple modes of design space exploration; they will be discussed
in detail in Section~\ref{sec:expl}.
The definition of these modes refers to the concept of
\emph{stability} of an assignment to system's parameters that satisfies all model constraints (which include the constraints defining the model itself and any constraint on model's interface).
We will refer to such a value assignment as a
\emph{stable witness}, or
\emph{(stable) solution} satisfying the model constraints. 
Informally, stability of a solution means that any eligible assignment in the specified region around the solution also satisfy the required constraints.
This notion is sometimes referred to as robustness.
We work with parameterized systems, where parameters (also called \emph{knobs}) can be tuned to optimize the system's performance under all legitimate inputs.
For example, in the circuit board design setting, topological layout of circuits, distances, wire thickness, properties of dielectric layers, etc.\ can be such parameters, and the exploration goal 
would be to optimize the system performance under the system's requirements~\cite{9501615}.
%
The difference between knobs and inputs is that knob values are selected during design phase, before the system goes into operation; on the other hand, inputs remain free and get values from the environment during the operation of the system. Knobs and inputs correspond to existentially quantified and universally quantified variables in the formal definition of model exploration tasks. Thus in the usual meaning of verification, optimization and synthesis, respectively, all variables are inputs, all variables are knobs, and some of the variables are knobs and the rest are inputs.

In this work by a
\emph{model}
we refer to an ML model that models the system under exploration.
The main capabilities of SMLP for system exploration include: 

\begin{description}
\item[assertion verification:]
    Verifying assertions on the model's interface.
\item[parameter synthesis:]
    Finding
    model
    parameter values such that 
    design constraints are valid.
\item[parameter optimization:]
    Optimizing the 
    model parameters under
    constraints.
\item[stable optimized synthesis:]
    Combining model parameter
    synthesis and
    optimization
    into one algorithm, enhanced by stability guarantees, to achieve safe, stable and %
    optimal 
    configurations.%
\item[root cause analysis:]
    Generating root-causing hints in terms of subset of parameters and their ranges that explain the failure.
\item[model refinement:]
    Targeted refinement of the model based on stability regions found by model exploration and on feedback from system in these stability regions.
\end{description}

\begin{figure}[bp]
\centering
\[
    \begin{tikzcd}[row sep=1.5em, column sep = {8.5em, between origins}, my label/.style={midway, sloped}]
    \bullet
    \arrow[rr, "\text{stable-optimization}"]
    \arrow[dr, swap,"\text{stable-synthesis}"{anchor=north,rotate=-18}]
    &&
    \bullet
    \arrow[dr]
    \\
    &
    \bullet
    \arrow[rr]
    &&
    \bullet
    \\
    \bullet
    \arrow[uu, "\text{stability}"{anchor=north,rotate=90}] 
    \arrow[rr, dashed, "\text{optimization}"] 
    \arrow[dr, swap, "\text{synthesis}"{anchor=south, rotate=-18}] 
    \arrow[rrru, dashdotted, "\text{stable optimized synthesis}"{anchor=south, rotate=6}]
    &&
    \bullet  
    \arrow[dr, dashed] 
    \arrow[uu, dashed]
    \\
    &
    \bullet
    \arrow[rr] 
    \arrow[uu]
    && 
    \bullet
    \arrow[uu]
    \end{tikzcd}
\]
\caption{Exploration Cube}
\label{fig:cube}
\end{figure}
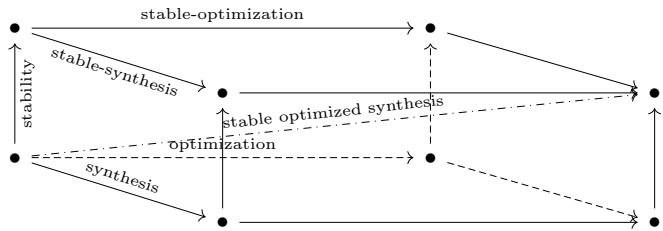

The \emph{model exploration cube} in Figure~\ref{fig:cube} provides a high level and intuitive idea on how the model exploration modes supported in SMLP are related.
The three dimensions in this cube represent synthesis ($\searrow$-axis), optimization ($\rightarrow$-axis) and stability ($\uparrow$-axis).
On the bottom plane of the cube, the edges represent the synthesis and optimization problems in the following sense:
synthesis with constraints configures the knob values in a way that guarantees that assertions are valid, but unlike optimization, does not guarantee optimally with respect to optimization objectives.
On the other hand, optimization by itself is not aware of assertions on inputs of the system and only guarantees optimality with respect to knobs, and not the validity of assertions in the configured system.
We refer to the process that combines synthesis with optimization and results in an optimal design that satisfies assertions as \emph{optimized synthesis}.
The upper plane of the cube represents introducing stability requirements into synthesis (and as a special case, into verification), optimization, and optimized synthesis.
The formulas that make definition of stable verification, optimization, synthesis and optimized synthesis precise are discussed in \changedto{Section~\ref{sect.exploration}}{Sections~\ref{sect.exploration} to \ref{sec:expl}}.


\delete{Will provide another version of cube where after optimization will use verification and after synthesis will use tuning (instead of optimization). Then need to comment that the cube is not confluent -- optimization + verification and synthesis + tuning does not always accomplishes the goal of optimized synthesis. \emph{FB: no need, the cube is illustrative, not a reference.}}%

\delete{
\begin{figure}
\centering
\[
    \begin{tikzcd}[row sep=1.5em, column sep = {8.5em, between origins}, my label/.style={midway, sloped}]
    \bullet
    \arrow[rr, "\text{stable-optimization}"]
    \arrow[dr, swap,"\text{stable-synthesis}"{anchor=north,rotate=-18}]
    &&
    \bullet
    \arrow[dr, swap, "\text{stable-synthesis}"{anchor=north,rotate=-14}]
    \\
    &
    \bullet
    \arrow[rr, "\text{stable-optimization}"]
    &&
    \bullet
    \\
    \bullet
    \arrow[uu, "\text{stability}"{anchor=north,rotate=90}] 
    \arrow[rr, dashed, "\text{optimization}"] 
    \arrow[dr, swap, "\text{synthesis}"{anchor=south, rotate=-18}] 
    \arrow[rrru, dotted, "\text{stable optimized synthesis}"{anchor=south, rotate=6}]
    &&
    \bullet  
    \arrow[dr, dashed, "\text{synthesis}"{anchor=north,rotate=-14}] 
    \arrow[uu, dashed, "\text{stability}"{anchor=north,rotate=90}]
    \\
    &
    \bullet
    \arrow[rr, "\text{optimization}"] 
    \arrow[uu, "\text{stability}"{anchor=north,rotate=90}]
    && 
    \bullet
    \arrow[uu, "\text{stability}"{anchor=north,rotate=90}]
    \end{tikzcd}
\]
\caption{Exploration Cube}
\label{fig:cube}
\end{figure}
} 

Compared to digital design, it is fair to say that formal methods have had a limited success in the analog domain. 
A practical approach to this challenge is to use models as a way of abstraction that can be refined based on model analysis and feedback from the real system to narrow the gap between the model and the system to levels tolerable by stability requirements of the design.
SMLP applies formal analysis to systems represented by ML models, and
assists 
designers in product development, as well as helps researchers to develop combined methods.
 
\delete{
\KK[inline]{ OLD:
One the other side of the spectrum are machine learning models and statistical methods, but they lack formal guarantees. 
 One practical approach to overcome this challenge is to combine formal methods with statistical and probabilistic approaches. 
and to use ML models as a way of abstraction that can be refined based on model analysis and feedback from the real system to narrow the gap between the model and the system to levels tolerable by stability requirements of the design.
SMLP combines statistical and probabilistic methods with formal analysis of systems represented by ML models, and can serve designers in product development, as well as help researchers to develop combined methods. 
}
}

\section{SMLP architecture}

\begin{figure}[tp]
\center
\includegraphics[width= 0.7\columnwidth]{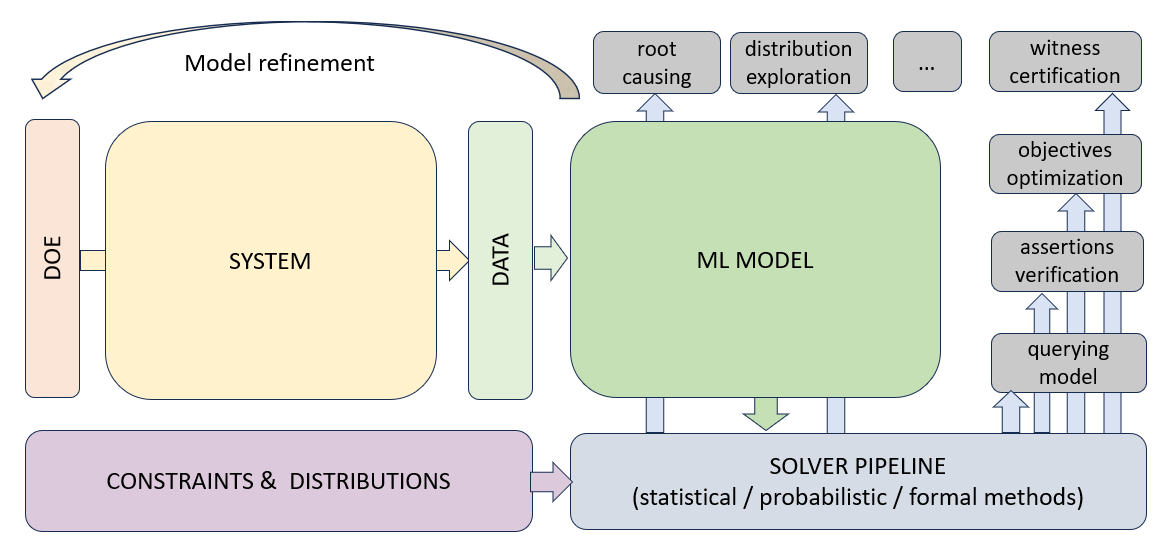}
\caption{SMLP Tool Architecture} \label{smlp_system}
\end{figure}

SMLP tool architecture is depicted in Figure~\ref{smlp_system}. 
It consists of the following components: 1) Design of experiments (DOE),
2) System that can be \changedto{simulated}{sampled} based on DOE, 3)  
ML model trained on the \changedto{simulated}{sampled} data, 4) SMLP solver that handles different system exploration modes on a symbolic representation of the ML model, 5) Targeted model refinement loop.

SMLP supports multiple ways to generate training data known under the name of \emph{Design Of Experiments (DOE).}
These methods include: full-factorial, fractional-factorial, Plackett-Burman, Box-Behnken, Box-Wilson, Sukharev-grid, Latin-hypercube, among other methods, which try to achieve a smart sampling of the entire input space with a relatively small number of data samples.
In Figure~\ref{smlp_system}, the leftmost box-shaped component called \textsc{doe} represents SMLP capabilities to generate test vectors to feed into the system and generate training data; the latter two components are represented with boxes called \textsc{system} and \textsc{data}, respectively. 

The component called \textsc{ml model} represents SMLP capabilities to train models; currently neural network, polynomial and tree-based regression models are supported. Modeling analog devices using polynomial models was proposed in the seminal work on \emph{Response Surface Methodology (RSM)}~\cite{51647df2-9154-322e-85ba-aa565c9e550a},
and since then has been widely adopted by the industry. Neural networks and tree-based models are used increasingly due to their wider adoption, and their exceptional accuracy and simplicity, respectively.

The component called \textsc{solver pipeline} represents model exploration engines of SMLP (e.g., connection to SMT solvers), which besides a symbolic representation of the model takes as input several types of constraints and input sampling distributions specified on the model's interface; these are represented by the component called \textsc{constraints \& distributions} located at the low-left corner of  Figure~\ref{smlp_system}, and
will be discussed in more detail in Section~\ref{sect.exploration}.
The remaining components represent the main model exploration capabilities of SMLP. 

Last but not least, the arrow connecting the \textsc{ml model} component back to the \textsc{doe} component represents a
\emph{model refinement} loop which allows to reduce the gap between the model and system responses in the input regions
where it matters for the task at hand (there is no need to achieve a perfect
match between the model and the system everywhere in the input space).
The targeted model refinement
loop is discussed in~\Cref{s.refinement}.

%

\section{Symbolic representation of models and constraints}\label{s.symbolic}
We assume that system interface consists of free inputs, knobs, and outputs.  The set of inputs and/or knobs, can be empty.
For the sake of ML-based analysis, we build an ML model,
represent it symbolically, and the aim is to analyze the system through exploring the model instead.

\delete{
\begin{figure}[!ht]
\center
\includegraphics[width= 0.4\columnwidth]{smlp_whitebox.PNG}
\caption{SMLP System Modeling} \label{smlp_whitebox}
\end{figure} \todokk{rm fig2}
}

A \emph{domain} $\mathcal D$ is a Cartesian product of reals, integers and finite non-empty sets.
A \emph{parameterized system} can be represented as a function
$f: \mathcal{D}_{\mathit{par}}\times \mathcal{D}_{\mathit{in}} \to
\mathcal{D}_{\mathit{out}}$,
where $\mathcal{D}_{\mathit{par}}, \mathcal{D}_{\mathit{in}}, \mathcal{D}_{\mathit{out}}$
are domains of parameters (knobs), inputs and outputs\changedto{}{,} respectively.
            \delete{%
\todozk{omit "e.g., reflecting interference between elements on a circuit board depending on the distance between elements as parameters.", it is not required and not so helpful; KK changed}
\KK[inline]{Can be removed or moved to intro: We also consider systems that are only implicitly represented.
Nevertheless, such systems can be sampled or simulated.
Based on such data we can build machine learning models which approximate the system and then analyze these models using symbolic methods.
We represent machine learning models using formulas in the language that can be handled by SMT solvers.}%
}%
For simplicity of the presentation we assume all domains are products of sets of reals but methods and implementation
are applicable
also for domains over integers and arbitrary finite sets.
We consider formulas over $\langle\mathbb R,0,1,\mathcal F,P\rangle$,
where $P$ contains the
usual predicates ${<},{\leq},{=}$, etc.\ and
$\mathcal F$ contains addition,
multiplication with rational constants and
can also contain non-linear functions supported by SMT solvers including polynomials, transcendental functions and
more generally computable functions~\cite{DBLP:conf/tacas/MouraB08,DBLP:conf/frocos/BrausseKKM19,DBLP:journals/tcs/BrausseKKM23,DBLP:conf/tacas/CimattiGSS13,DBLP:conf/cade/GaoAC12}.
We extend functions $\mathcal F$ by functions 
\emph{definable} by formulas: $\mathcal F_D$, i.e., we assume $f\in \mathcal{F_D}$ is represented by a
formula $F(x_1,\ldots,x_n,y)$ over variables $x_1,\ldots,x_n$ corresponding to the $n$ inputs and $y$ corresponding to the output $f(x_1,\ldots,x_n)$.
We assume that satisfiability of quantifier free formulas over this language is decidable or more generally $\delta$-decidable~\cite{DBLP:conf/cade/GaoAC12,DBLP:journals/tcs/BrausseKKM23}. 
Let us note that even when basic functions $\mathcal F$ contain just linear functions,
$\mathcal F_D$ will contain, e.g.,
functions represented by neural networks with $\operatorname{ReLU}$ activation functions as well as 
decision trees and random forests.  
When representing parameterized systems using ML models we assume that parameters are treated as designated inputs to the ML model.

Throughout, $p,x,y$ denote respectively knob, input and output 
variables (or variable vectors) in formulas while $r,z$ range over reals.
Whenever we use a norm $\Vert\cdot\Vert$, we refer to a norm representable in our language, such as the Chebyshev norm $(x_1,\ldots,x_n)\mapsto\max\{|x_1|,\ldots,|x_n|\}$. 

\delete{
Figure~\ref{smlp_model} depicts a high-level architecture for generating a symbolic representation of ML models.
\begin{figure}[!ht]
\center
\includegraphics[width= 0.7\columnwidth]{smlp_model.PNG}
\caption{SMLP Model Exploration} \label{smlp_model}
\end{figure}
}

\section{Symbolic representation of the ML model exploration}\label{sect.exploration}



The main system exploration tasks handled by SMLP 
can be defined using the GEAR-fragment of $\exists^*\forall^*$ formulas~\cite{DBLP:conf/fmcad/BrausseKK20}:
\begin{equation}\label{form:gear:final}
    \exists p ~\added{\big[}\eta(p) \wedge
    \forall p'~
    \forall x y~[
    \theta(p,p') \implies (\varphi_M(p',x,y)  \implies  \varphi_{\mathit{cond}}(p',x,y))
    ]\added{\big]}
\end{equation}
where 
$x$ ranges over inputs, $y$ ranges over outputs, and
$p,p'$ range over knobs,
 $\eta(p)$ are constraints on the knob configuration $p$,
  $\varphi_{M}(p',x,y)$ defines the machine learning model, $\theta(p,p')$ defines stability region for the solution $p$, and $\varphi_{\mathit{cond}}(p',x,y)$ defines conditions that should hold in the stability region.
An assignment to variables $p$ that makes formula \eqref{form:gear:final} true is called a \emph{$\theta$-stable witness}, or simply a \emph{solution}, to~\eqref{form:gear:final}. 


In our formalization $\theta,\eta$ and $\varphi_{\mathit{cond}}$ are quantifier free formulas in the language. These constraints and how they are implemented in SMLP are described below.  


\begin{description}
\item[$\eta(p)$]
    Constraints on values of knobs; 
    this formula need not be a conjunction of constraints on individual knobs, can define more complex relations between allowed knob values of individual knobs. $\eta(p)$ can be specified through the SMLP specification file (see \Cref{sec:spec-file})
    . 
\item[$\theta(p,p')$]
    Stability constraints that define a region around a candidate solution. 
    This can be specified using either absolute or relative radius $r$ in the 
    specification
    file.
    This region corresponds to a ball (or box) around $p$: $\theta(p,p')\eqdef \Vert p - p' \Vert \leq r$. In general, our methods do not impose any restrictions on $\theta$ apart from reflexivity.
\item[$\varphi_M(p,x,y)$]
    Constraints that define the function represented by the ML model $M$, thus
    $\varphi_M(p,x,y) \eqdef(M(p,x) = y)$.  
    In the ML model\changedto{}{,} knobs are represented as designated inputs (and can be treated in the same way as system inputs, or the machine model architecture can reflect the difference between inputs and knobs).
    $\varphi_M(p,x,y)$ is computed by SMLP internally, based on the ML model specification.
\item[$\varphi_{\mathit{cond}}(p,x,y)$]    
    Conditions that should hold in the $\theta$-region of the  solution.
    These conditions depend on the exploration mode and could be:
    (1) verification conditions,
    (2) model querying conditions, 
    (3) parameter optimization conditions, or
    (4) parameter synthesis conditions.
    The exploration modes are described in 
    \Cref{sec:expl}.
\end{description}

SMLP solver 
is based on specialized procedures for solving formulas in the GEAR fragment using quantifier-free SMT solvers, GearSAT$_\delta$~\cite{DBLP:conf/fmcad/BrausseKK20} and GearSAT$_\delta$-BO~\cite{DBLP:conf/ijcai/BrausseKK22}.
The GearSAT$_\delta$ procedure interleaves search for candidate solutions using SMT solvers with exclusion of $\theta$-regions around counterexamples. GearSAT$_\delta$-BO combines GearSAT$_\delta$ search with Bayesian optimization guidance.
These procedures find
solutions to GEAR formulas
with user-defined accuracy $\varepsilon$ (
defined in \Cref{sec:opt-synth}) and they have been proven to be sound, ($\delta$)-complete and terminating%
.

\section{ 
Problem specification in SMLP}
\label{sec:spec-file}
\begin{figure}[tp]
\scriptsize
\begin{verbatim}
{
  "version": "1.2",
  "variables": [
    {"label":"y1", "interface":"output", "type":"real"},
    {"label":"y2", "interface":"output", "type":"real"},
    {"label":"x1", "interface":"input", "type":"real", "range":[0,10]},
    {"label":"x2", "interface":"input", "type":"int", "range":[-1,1]},
    {"label":"p1", "interface":"knob", "type":"real", "range":[0,10], "rad-rel":0.1, "grid":[2,4,7]},
    {"label":"p2", "interface":"knob", "type":"int", "range":[3,7], "rad-abs":0.2}
  ],
  "alpha": "p2<5 and x1==10 and x2<12",
  "beta": "y1>=4 and y2==8",
  "eta": "p1==4 or (p1==8 and p2 > 3)",
  "assertions": {
    "assert1": "(y2**3+p2)/2>6",
    "assert2": "y1>=0",
    "assert3": "y2>0"
  },
  "objectives": {
    "objective1": "(y1+y2)/2",
    "objective2": "y1"
  }
}
\end{verbatim}
\begin{tikzpicture}[remember picture,overlay,shift={(22em,\baselineskip)}]
\node[anchor=south west]{\includegraphics[height=14\baselineskip]{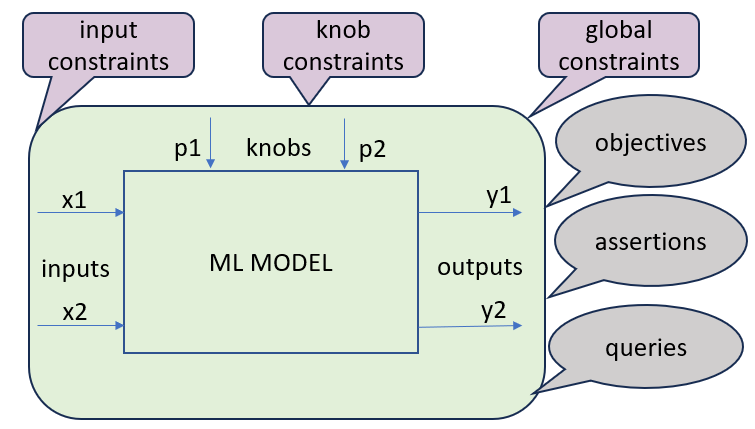}};
\end{tikzpicture}
\vspace*{-1\baselineskip}
\caption{Example of SMLP's format specifying the problem conditions for the displayed model of the system.}
\label{fig:spec}
\end{figure}
The specification file 
defines the problem conditions in a JSON compatible format, whereas SMLP exploration modes can be specified via command line options.
\Cref{fig:spec} depicts a toy system with two inputs, two knobs, and two outputs and a matching specification file for model exploration modes in SMLP.
For each variable it specifies its \emph{label} (the name),
its \emph{interface} function (``input'', ``knob'', or ``output''),
its \emph{type} (``real'', ``int'', or ``set'', for categorical features),
\emph{ranges} for variables of real and int types, and optionally,
a \emph{grid} of values for knobs that they are allowed to take on within the respective declared ranges, independently from each other (unless there are constraints further restricting the multi-dimensional grid). Both integer and real typed knobs can be restricted to grids (but do not need to).
\delete{
\KK[inline]{ MOVE somewhere SMLP problem specification is done in a JSON compatible format that is submitted FB: Missing: alpha,
beta, eta, assertions,
objectives
to SMLP via a specification file.else: Some of the models, such as neural networks and polynomial models, require categorical variables to be encoded as numeric features, say using one-hot encoding or similar, while tree-based models can train with categorical features directly.} %
}%
%
Additional fields \emph{alpha}, \emph{beta}, \emph{eta}, \emph{assertions} and \emph{objectives} can optionally be specified, as shown in the example.
These correspond to the predicates $\alpha$, $\beta$, $\eta$, `$\operatorname{assert}$' and
objective function $\objv$
described in
\Cref{sec:expl}.
The details about the concrete format are described in the manual distributed with SMLP in the \texttt{doc/spec} subdirectory.%
\todofb{PDF version in repo; rename main branch}%

\section{SMLP exploration modes of ML models}\label{sec:expl}

In this section we describe ML model exploration modes supported by SMLP,
which are based on Formula~\eqref{form:gear:final}.

\subsection{Stable parameter synthesis}\label{subsec:find}
The goal of \emph{stable synthesis} is to find values of the system parameters 
such that 
\changedto{verification}{required} conditions hold in the $\theta$-region of the parameters for all inputs. 
For this, SMLP solves Formula~\eqref{form:gear:final}, where 
\[\varphi_{\mathit{cond}}(p,x,y)\eqdef \alpha(p,x) \implies \beta(p,x,y)\text.\]
Here, $\alpha(p,x)$ restricts points in the region around the solutions to  points of interest and $\beta(p,x,y)$ is the requirement that these points should satisfy.
The $\alpha$ constraints define the domain of inputs and knobs and constraints on them which play the role of assumptions in the assume-guarantee paradigm,
while $\beta$ constraints can be viewed as guarantees; 
they can express some external/additional requirements from system not covered by assertions.
In case of synthesis and optimization, $\beta$ constraints can be used to express constraints that should be satisfied by synthesized, respectively,  optimized system. 
For example consider $\alpha(p,x) \eqdef (x_1 > x_2 + x_3)$, 
$\beta(p,x,y) \eqdef y_1 > 2\cdot x_1$ and $\theta\eqdef\Vert p - p'\Vert \leq 0.5$. 
In this mode SMLP will find value of parameters of the system such that for all parameters
in the $0.5$ region and all inputs such that $x_1 > x_2 + x_3$ the output value $y_1$ is greater than $2\cdot x_1$.

\subsection{Querying conditions on the model}
As mentioned in \Cref{sect.exploration},
the methods behind SMLP are ($\delta$)-complete, 
hence SMLP can also
determine that a solution does not exist.
In these cases one usually wants to relax $\beta$ condition into $\beta'$ and query the model on them.
This can be done using:
\[\varphi_{\mathit{cond}}(p,x,y) \eqdef \alpha(p,x) \implies \beta'(p,x,y) \wedge \query(p,x,y).\]
where $\query(p,x,y)$ are conditions of interest. Relaxation and strengthening of $\beta$ and queries can be done iteratively, and is a way to express and solve soft constraints. 
In SMLP soft constraints can also be solved by encoding them as an optimization problem, described in~\Cref{sec:opt-synth}.

\subsection{Verifying assertions on a model}

For verifying an assertion $\assert(p,x,y)$ on a model $M$ under given parameters $p$ we can simplify 
Formula~\eqref{form:gear:final} to: 
\[
    \eta(p) \wedge \forall p' ~\forall x y ~[
    \theta(p,p') \implies (\varphi_M(p',x,y)  \implies \assert(p',x,y))
    ]\text.
\]
Since $p$ is fixed, $\eta(p)$ can be eliminated by evaluation.
Further, if one is not concerned with stability, then $\theta$ can be replaced with the identity and the problem can be reduced to a standard verification problem. 
\[
     ~\forall x y~(\varphi_M(p,x,y)  \implies \assert(p,x,y))\text.
\]

In the case of neural networks,
there is a large range of verification tools to solve this problem such as Marabou~\cite{DBLP:conf/cav/KatzHIJLLSTWZDK19}, $\alpha$,$\beta$-CROWN~\cite{xu2021fast,wang2021beta}. 
Most of these tools rely on floating point computations, which can quickly accumulate errors.
SMLP supports SMT solvers with arbitrary precision which can produce exact results, at the expense of the computational cost. Nevertheless, dedicated ML solvers are very useful as they scale to much larger problems~\cite{DBLP:journals/sttt/BrixMBJL23}. We are currently working on supporting dedicated ML solvers in SMLP and let user decide which traded-off to choose.
SMLP also supports other ML models such as decision trees, random forests and  polynomial models.

\subsection{Stable optimized synthesis}
\label{sec:opt-synth}
\begin{figure}[tp]
\center
\includegraphics[width= 0.7\columnwidth]{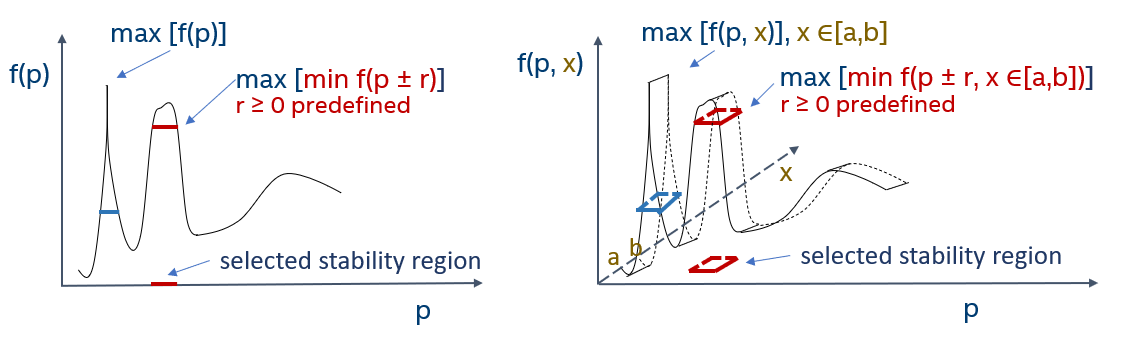}
\caption{SMLP max-min optimization. On both plots, $p$ denote the knobs. On the right plot we also consider inputs $x$ (which are universally quantified) as part of $f$.} \label{smlp_maxmin}
\end{figure}

In this subsection we consider the optimization problem for a real-valued function $f$ (in our case, an ML model),
extended in two ways:
(1) we consider a $\theta$-stable maximum to ensure that the objective function  does not drop drastically in a close neighborhood of the configuration where its maximum is achieved, and
(2) we assume that the objective function besides knobs depends also on inputs, and the 
function is maximized in the stability $\theta$-region of knobs, for any values of inputs in their respective legal ranges.
We explain these extensions using two plots in Figure~\ref{smlp_maxmin}.

The left plot represents optimization problem for $f(p,x)$ when $f$ depends on knobs only (thus $x$ is an empty vector), while the right plot represents the general setting where $x$ is not empty (which is usually not considered in optimization research). In each plot, the blue threshold (in the form of a horizontal bar or a rectangle) denotes the stable maximum around the point where $f$ reaches its (regular) maximum, and the red threshold denotes the stable maximum, which is approximated by our optimization algorithms. In both plots, the regular maximum of $f$ is not stable due to a sharp drop of $f$'s value in the stability region. 
\delete{The right plot depicts the situation when input $x$ ranges in interval $[a, b]$, thus in \emph{max-min} optimization problem, formalized below in Formulas~\eqref{form:opt1}  and  \eqref{form:opt2}, the minimum of $f$ is calculated in the stability region of knobs $p$, with values of $x$ ranging in $[a,b]$.}

%
%
%
%

\strut\todozk{remove `near' everywhere}%
Let us first consider optimization without stability or inputs, i.e., far low corner in the exploration cube Figure~\ref{fig:cube}. 
Given a 
formula
$\varphi_M$ encoding the model, and an objective function 
$\objv:%
\mathcal{D}_{\mathit{par}}\times
\mathcal{D}_\mathit{out}\to\mathbb R$, the
standard optimization problem solved by SMLP is stated by Formula~\eqref{form:opt_p}.
 \begin{equation}\label{form:opt_p}
 \regmax{\varphi_M}{\objv}
 \eqdef\mathop{\max\limits_{p}} \{z \mid  
    \forall y\changedto[{~(} \varphi_M(p,y)  \implies  
    \objv(p,y) \geq z
    \changedto])\}
\end{equation}
A solution to this optimization problem 
is the pair $(p^*,\regmax{\varphi_M}{\objv})$, where 
$p^*{}\in\mathcal{D}_{\mathit{par}}$ is a value of parameters $p$ on which the maximum $\regmax{\varphi_M}{\objv}\in\mathbb R$ of the objective function $\objv$ is achieved
for the output $y$ of the model on $p^*$.
In most cases it is not feasible to exactly compute the maximum.
To deal with this, SMLP computes maximum with a specified accuracy. 
Consider $\varepsilon >0$. 
We refer to 
values $(\tilde p,\tilde z)$ as a
solution to the optimization problem with
\emph{accuracy} $\varepsilon$, or \emph{$\varepsilon$-solution},
if $\tilde z\leq \regmax{ \varphi_M}{\objv}  <\tilde z+\varepsilon$ holds and
%
$\tilde z$ is a lower bound on the objective, i.e., $\forall y [ \varphi_M(p,y)  \implies  
    \objv(p,y) \geq \tilde z] $ holds.

Now, we consider \emph{stable optimized synthesis}, i.e., the top right corner of the exploration cube.
The problem can be formulated as the following Formula~\eqref{form:opt1}, expressing maximization of a lower bound  on the objective function $\objv$ over parameter values under stable synthesis constraints.
\begin{equation}\label{form:opt1}
\regmax{\varphi_M}{\objv,\theta}
\eqdef\mathop{\max\limits_{p}} \{z \mid\eta(p) \wedge
    \forall p'~
    \forall x y~[
    \theta(p,p') \implies
    (\varphi_M(p',x,y)  \implies 
     \varphi_{\mathit{cond}}^{\geq}(p',x,y,z))
    ]\}
\end{equation}
where
\[
\varphi_{\mathit{cond}}^{\geq}(p',x,y,z)
\eqdef \alpha(p',x) \implies (\beta(p',x,y) \wedge \objv(p',x,y) \geq z).
\]
The stable synthesis constraints are part of a GEAR formula and include usual $\eta, \alpha, \beta$ constraints together with the stability constraints $\theta$.
%
%
Equivalently, stable optimized synthesis can be stated as the \emph{max-min} optimization problem, Formula~\eqref{form:opt2}
\begin{equation}\label{form:opt2}
\regmax{\varphi_M}{\objv,\theta}
\eqdef
\mathop{\max\limits_{p}} \mathop{\min\limits_{x, p'}}
\{ z \mid \eta(p) \wedge
    \forall y~[
     \theta(p,p') \implies
     (\varphi_M(p',x,y)  \implies  \varphi_{\mathit{cond}}^{\leq}(p',x,y,z))]\}
\end{equation}
%
%
where \[
\varphi_{\mathit{cond}}^{\leq}(p',x,y,z)
\eqdef\alpha(p',x) \implies (\beta(p',x,y) \wedge \objv(p',x,y) \leq z)\text.\]
In Formula~\eqref{form:opt2} the minimization predicate in the stability region corresponds to the universally quantified $p'$ ranging over this region in~\eqref{form:opt1}.
An advantage of this formulation is that this formula can be adapted to define other aggregation functions over the objective's values on stability region.
For example, that way one can represent the \emph{max-mean} optimization problem, where one wants to maximize the mean value of the function in the stability region rather one the min value (which is maximizing the worst-case value of $f$ in stability region).
Likewise, Formula~\eqref{form:opt2} can be adapted to other interesting statistical properties of distribution of values of $f$ in the stability region.

We can explicitly incorporate assertions in stable optimized synthesis by defining $\beta(p',x,y)\eqdef \beta'(p',x,y) \wedge \assert(p',x,y)$, where  $\assert(p',x,y)$ are assertions required to be valid in the entire stability region around the selected configuration of knobs $p$.
The notion of $\varepsilon$-solutions for these problems carries over from the one given above for Formula~\eqref{form:opt_p}.

SMLP implements stable optimized synthesis based on the GearOPT$_\delta$ and GearOPT$_\delta$-BO algorithms~\cite{DBLP:conf/fmcad/BrausseKK20,DBLP:conf/ijcai/BrausseKK22}, which are shown to be complete and terminating 
for this problem under mild conditions. 
%
These algorithms were further extended in SMLP to Pareto point computations to handle multiple objectives simultaneously.


\subsection{Design of experiments}\label{sssect:doe}

Most DOE methods
are 
based on understanding multivariate distribution of legal value combinations
of inputs and knobs 
in order to sample the system.
\pagebreak[3]
When the number of system inputs and/or knobs is large (say hundreds or more), the DOE may not  generate a high-quality coverage of the system’s behavior to enable training models with high accuracy.
Model training process itself becomes less manageable when number of input variables grows,
and models are not explainable and thus cannot be trusted.
One way to curb this problem is to select a subset of input features for DOE and for model training.
The problem of combining feature selection with DOE generation and model training
is an important research topic of practical interest,
and SMLP supports multiple practically proven ways
to select subsets of features and feature combinations as inputs to DOE and training,
including the \emph{MRMR} feature selection algorithm~\cite{DBLP:journals/jbcb/DingP05},
and a \emph{Subgroup Discovery (SD)} algorithm~\cite{DBLP:books/mit/fayyadPSU96/Klosgen96,DBLP:conf/pkdd/Wrobel97,DBLP:journals/widm/Atzmueller15}.
The MRMR algorithm selects a subset of features according to the principle of \emph{maximum relevance and minimum redundancy}.
It is widely used for the purpose of selecting a subset of features for building accurate models,
and is therefore useful for selecting a subset of features to be used in DOE;
it is a default choice in SMLP for that usage.
The SD algorithm selects regions in the input space relevant to the response,
using heuristic statistical methods,
and such regions can be prioritized for sampling in DOE algorithms.

\subsection{Root cause analysis}

We view the problem of root cause analysis as dual to the stable optimized synthesis problem: while during optimization with stability we are searching for regions in the input space (or in other words, characterizing those regions) where the system response is good or excellent, the task of root-causing can be seen as searching for regions in the input space where the system response is not good (is unacceptable). Thus simply by swapping the definition of excellent vs unacceptable, we can apply SMLP to explore weaknesses and failing behaviors of the system.

Even if a number of witnesses (counter-examples to an assertion) are available, they represent discrete points in the input space and it is not immediately clear which value assignments to which variables in these witnesses are critical to explain the failures. Root causing capability in SMLP is currently supported through two independent approaches:
a \emph{Subgroup Discovery (SD)} algorithm that searches through the data for the input regions where there is a higher ratio (thus, higher probability) of failure; to be precise, SD algorithms support a variety of \emph{quality functions} which play the role of optimization objectives in the context of optimization.
To find input regions with high probability of failure, SMLP searches for stable witnesses to failures. These capabilities, together with feature selection algorithms supported in SMLP, enable researchers to develop new root causing capabilities that combine formal methods with statistical methods for root cause analysis.

\subsection{Model refinement loop}\label{s.refinement}

Support in SMLP for selecting DOE vectors to sample the system and generate a training set was discussed in Subsection~\ref{sssect:doe}.
Initially, when selecting sampling points for the system, it is unknown which regions in the input space are really relevant for the exploration task at hand.
Therefore 
some DOE algorithms also incorporate
sampling based on previous experience and familiarity with the design, such as sampling nominal cases and corner cases, when these are known.
For model exploration tasks supported by SMLP, it is not required to train a model that will be an accurate match to the system everywhere in the legal search space of inputs and knobs.
We require to train a model that is an \emph{adequate} representation of the system for the task at hand, meaning that the exploration task solved on the model solves this task for the system as well.
Therefore SMLP supports a \emph{targeted model refinement} loop to
enable solving the system exploration tasks by solving these tasks on the model instead.
The idea is as follows:
when a stable solution to model exploration task is found, it is usually the case that there are not many training data points close to the stability region of that solution.
This implies that there is a high likelihood that the model does not accurately represent the system in the stability region of the solution.
Therefore the system is sampled in the stability region of the solution, and these data samples are added to the initial training data to retrain the model and make it more 
adequate in the stability region of interest.
\todo{Samples in the stability region
of interest can also be assigned higher weights compared to other samples to help training to achieve higher accuracy in that region. More generally, higher adequacy of the model can be achieved by sampling distributions biased towards prioritizing the stability region during model refinement.}

\section{Implementation}

\strut\KK{refer to SMLP manual on details how to train ML models, specify constraints, modes etc.}%
SMLP code is open-source and publicly available%
\footnote{\url{https://github.com/fbrausse/smlp}}.
Its frontend is implemented in Python, and its backend is implemented in C++
,
while the interface between the two is realized using the Boost library.
For training tree-based and polynomial models we use the scikit-learn%
\delete{\footnote{\url{https://scikit-learn.org/stable/}}} and pycaret%
\delete{\footnote{\url{https://pycaret.org}}} packages, and for training neural networks we use the Keras package with TensorFlow%
\delete{\footnote{\url{https://keras.io}}}.
Our focus is on analyzing regression models arising from systems with analog pins and analog output,
but classification models are also covered as they can be reduced to binary classification
with output values $0$ and $1$,
or by treating the binary classification 
problem as a regression problem
of predicting the probability of the output to be $1$
(the latter is usually preferable for more finer analysis).
For generating training data from a system,
SMLP supports DOE approaches available in package
pyDOE. 
The MRMR algorithm for feature selection is integrated in SMLP using the mrmr package, and the Subgroup Discovery algorithm is integrated using package pysubgroup.

SMLP can use any external SMT solver which supports SMT-LIB2 format, as a back end of the GearSAT/OPT algorithms (via command line options), and also natively integrates Z3 via the Python interface.
We \todozk{We have??? \emph{FB: experimented, yes :)}}%
successfully experimented with Z3~\cite{DBLP:conf/tacas/MouraB08}, Yices~\cite{DBLP:conf/cav/Dutertre14}, CVC5~\cite{DBLP:conf/tacas/BarbosaBBKLMMMN22}, MathSAT~\cite{DBLP:conf/tacas/CimattiGSS13} and \texttt{ksmt}
~\cite{DBLP:conf/frocos/BrausseKKM19}.


\section{Experimental results}

Previous publications~\cite{DBLP:conf/fmcad/BrausseKK20,DBLP:conf/ijcai/BrausseKK22}
on SMLP report detailed experimental results on $10$ real-life training datasets originating form Electrical Validation and Signal Integrity domains.
The output is an analog signal measuring the quality of a transmitter or a receiver of a channel to a peripheral device.
The datasets are freely available\footnotemark[3]
: 
$5$ transmitter (TX) datasets and $5$ receiver (RX) counterparts.
The count of inputs and knobs together in these experiments, as well as in current usage of the SMLP tool at Intel, is around $5$ to $20$ variables.
In~\cite{DBLP:conf/fmcad/BrausseKK20} the experimental evaluation is performed using \textsc{GearSat$_\delta$} algorithm, and experimental results using the \textsc{GearOpt$_\delta$-BO} algorithm that combines SMT-based optimization procedure with Bayesian optimization~are reported in \cite{DBLP:conf/ijcai/BrausseKK22}.
While these datasets are relatively small in terms of parameter counts, they are
representative of modeling IO devices at Intel, and SMLP has been useful in suggesting safe and
optimized configurations for a number of real-life IO devices in recent years. \todozk{Drop ``near'' or define it after $\epsilon$-accurate solution was defined.}
\KK[inline]{Maybe add: Note that already functions with $2$ inputs can be challenging for state-of-the-art methods, e.g., Rastrigin’s Function, Griewank’s function, Ackley’s function etc. Hence, the number of inputs does not necessarily correlate with the complexity of the function.
In the applications that we considered so far, the number of input variables is $5-20$, but the functions are generally more amenable to analysis. These problems are coming from industrial application and SMLP demonstrated its feasibility and usefulness for \changed{analog device} optimization at Intel. Note that scalability of SMLP solely depends on scalability of the background SMT or ML solvers, hence we believe that SMLP will scale together with SMT and ML solvers.
} \todozk{There is some repetition in this comment compared to preceding paragraph. Remove this repetition. Also, can we change 'solily' to 'largly'? \emph{FB: `mainly'?}}

\section{Future Work}

Currently we are extending SMLP to support ONNX 
format used by VNN-LIB
~\cite{demarchi2023supporting} so more specialized solvers for ML can be also used alongside SMT solvers.
We are working on combining
different solving strategies into a user-definable solver pipeline
of ML and SMT solvers within the SMLP framework. 
We are planning to
release more real-life industrial datasets 
in near future.

\todozk{More statistical approaches will be enabled? 
Solver pipeline?}

\bibliography{bib}


\end{document}